\def\year{2021}\relax
\documentclass[letterpaper]{article} 
\usepackage{aaai21}  
\usepackage{times}  
\usepackage{helvet} 
\usepackage{courier}  
\usepackage[hyphens]{url}  
\usepackage{graphicx} 
\def\UrlFont{\rm}  
\usepackage{breqn}
\usepackage{natbib}  
\usepackage{caption}
\usepackage{amsmath}
\usepackage{amssymb}
\usepackage{booktabs}
\usepackage[switch]{lineno}  %
\usepackage{siunitx}
\usepackage[colorinlistoftodos]{todonotes}

\graphicspath{ {./images_low_res/} }
\title{Learning Potentials of Quantum Systems using Deep Neural Networks}
\author{Arijit Sehanobish,\thanks{Equal Contribution}\textsuperscript{\rm 1} Hector H. Corzo,\footnotemark[1]\textsuperscript{\rm 2} Onur Kara,\textsuperscript{\rm 3} David van Dijk\textsuperscript{\rm 1} \\} 
\affiliations {
\textsuperscript{\rm 1} Internal Medicine (Cardiology) and Computer Science, Yale University \\ \texttt{\{arijit.sehanobish, david.vandijk\}@yale.edu} \\
\textsuperscript{\rm 2} Center for Chemical Computation and Theory \\
University of California, Merced\\
\texttt{hhcorzo@gmail.com}\\ 
\textsuperscript{\rm 3} Hindsight Technology Solutions\\
\texttt{okara83@gmail.com}
}

\begin{document}
\maketitle

\begin{abstract}
Attempts to apply Neural Networks (NN) to a wide range of research problems have been ubiquitous and plentiful in recent literature. Particularly, the use of deep NNs for understanding complex physical and chemical phenomena has opened a new niche of science where the analysis tools from  Machine Learning (ML) are combined with the computational concepts of  the natural sciences. Reports from this unification of ML have presented evidence that NNs can learn classical Hamiltonian mechanics. This application of NNs to classical physics and its results motivate the following question: Can NNs be endowed with inductive biases through observation as means to provide insights into quantum phenomena? In this work, this question is addressed by investigating possible approximations for reconstructing the Hamiltonian of a quantum system in an unsupervised manner by using only limited information obtained from the system's probability distribution.


\end{abstract}

\section{Introduction}
 In the Machine Learning (ML) realm, Neural Networks (NNs) are among the most used and exceptionally efficient models to learn and generalize information from data.
These data interpretative capabilities have provoked the widespread use of NNs in Natural Language Processing~\cite{torfi2020natural}, Image Classification~\cite{alex2019big}, Video Captioning~\cite{sun2019videobert} and Reinforcement Learning~\cite{du2019taskagnostic, higgins2016early}; furthermore, recent works have shown the capabilities of NNs in symbolic reasoning and  mathematical problem solving~\cite{lample2019deep}. 
 In the case of natural sciences, applying ML to physics is not new,  several works have been reported~\cite{toth2019hamiltonian,greydanus2019hamiltonian,cranmer2020lagrangian,tong2020symplectic} where different authors have combined ML with Hamilton's equations of motion to generate trajectories that obey energy conservation principles and  classical physical laws. In material sciences, on the other hand,  ML has proven to be an important interpretative tool for the computational prediction of new materials~\cite{schleder2019dft}. The encouraging results reported by the different applications of ML have motivated the use of NNs as powerful tools to gain insight into the laws of physics that govern the behavior of complicated natural classical  phenomena.  
Unlike classical physics, in quantum physics, objects have characteristics of both particles and waves (wave--particle duality) for which the concept of trajectory is no longer defined nor can their position and momentum, both, be measured simultaneously~\cite{Sakurai,robinett,feynman1965feynman,robinett2006}.
Because of this wave--particle duality, the state of a quantum mechanical system is fully specified by its wave--function, which is typically obtained by solving the Schr\"{o}dinger equation~\cite{Sakurai,robinett,feynman1965feynman,robinett2006}. In many cases, however, not only solving this equation is difficult, but also its correct formulation requires knowledge about the form of the potential energy operator, 
 which often may not be completely known. In contrast, the inverse form of the Schr\"{o}dinger equation~\cite{nakatsuji2002inverse,chadan2012inverse,zakhariev2012direct,jensen2018numerical} presents an alternative for describing quantum phenomena by reformulating the description of quantum mechanical systems as solutions of inverse problems~\cite{aster2018parameter,groetsch1993inverse,vogel2002computational}. Inverse problems are central to the study of quantum mechanical systems, insomuch that much of what is known about the electronic structure of matter has been mathematically characterized by solutions of inverse problems~\cite{beals2009strings,zakhariev2012direct,jensen2018numerical,vogel2002computational}. Thus,  numerical algorithms for the inversion of the Schr\"{o}dinger equation are important predictive tools for the further development of approximate quantum mechanical methodologies such as scattering approximated models~\cite{zakhariev2012direct}, Density Functional Theory (DFT)~\cite{jensen2018numerical}, etc. 


In this work, rather than handcrafting numerical solutions for the inverse Schr\"{o}dinger equation~\cite{aster2018parameter,jensen2018numerical,vogel2002computational,beals2009strings} to define a potential function and describe a quantum phenomena, a neural network, termed Quantum Potential Neural Network (QPNN), is instead designed to learn potential functions directly from observables in an unsupervised manner. This proposed QPNN for learning potential functions was developed based on the underlying formalism for the inverse solution of the  Schr\"{o}dinger equation.
Thus, the proposed QPNN opens the possibility for generating simpler and succinct functions that can be used to construct effective Hamiltonians for the description of a variety of quantum systems using only a small portion of the available information known about the system. Since these effective Hamiltonians can be generalized to obtain other observables, QPNN  may provide unique insights into complex quantum phenomena were only a small amount of information is available.

\section{Theory}
The mathematical description of a quantum particle typically takes the form of a complex function of spatial coordinates $\Vec{x}$ and time coordinates $t$ called wave--function, $\Psi(\Vec{x},t)$ ~\cite{Sakurai,robinett,feynman1965feynman,robinett2006}.
$\Psi(\Vec{x},t)$ is a complex--valued probability amplitude whose square modulus  ($|\Psi(\Vec{x},t)|^{2}$) correspond to  the probability of finding the particle described by  the wave--function at that given $\Vec{x}$ and $t$.
The classically measured value of a physical observable, however, is not given directly by  $\Psi(\Vec{x},t)$  but by the expectation values of the operators that represent the desired measurement  acting on $\Psi(\Vec{x},t)$. 
%
In many scenarios, wave--functions are obtained as direct solutions of the time--dependent Schr\"{o}dinger equation,  
\begin{equation}\label{eqn:TDSE}
i\hbar\frac{\partial \Psi(\Vec{x},t)}{\partial t} = \hat{\rm H} \Psi(\Vec{x},t),
\end{equation}
where $\hbar$ is Planck's constant 
and $\hat{\rm H}$ is the Hamiltonian operator of the system, which is an Hermitian operator acting on an infinite dimensional space of $L^{2}$ functions. Thus, $\hat{\rm H}$ needs not be compact and as much may not have any eigenvalues. When $\hat{\rm H}$ is time--independent, equation \ref{eqn:TDSE} can be reduced to the following equation
\begin{equation}\label{eqn:TISE}
    \hat{\rm H}\psi_{n}(\Vec{x})=E\psi_{n}(\Vec{x}),
\end{equation}
where $n$ indicates the quantum state of the system.
In many cases, the physical information contained in the time--independent wave--function $\psi_{n}(\Vec{x})$ may be enough for the characterization of the system under study.   

%
\subsection{Hamiltonian}
The Hamiltonian  operator, $\hat{\rm H}$, is fundamental in many formulations of quantum theory. This operator is expressed as the sum of the kinetic ($\hat{\rm T}$) and  potential energy operators ($\hat{\rm V}$) for all particles in the quantum system,
\begin{equation}\label{eqn:Hamilt}
    \hat{\rm H}=\hat{\rm T}+\hat{\rm V}.
\end{equation}
Generally, the kinetic energy operator contained in $\hat{\rm H}$ only depends on the second derivatives of the wave--function, with respect to its spatial coordinates. The potential energy operator, however, depends on the physical circumstances imposed onto the system, and varies from system to system. Thus equation~\ref{eqn:Hamilt} may be expressed as 
\begin{equation}
    \hat{\rm H}= -\frac{\hbar^{2}}{2m}\frac{\partial^2}{\partial \Vec{x}^{2}}+\hat{\rm V}(\Vec{x},t)\equiv -\frac{\hbar^{2}}{2m}\nabla_{\Vec{x}}^{2}+\hat{\rm V}(\Vec{x},t). 
\end{equation}
In quantum mechanics, the problem of finding the $\hat{\rm H}$ that characterizes a given phenomenon could be reduced to formulating the potential operator that contains all the governing physical descriptors.

\begin{figure*}[t]
    \centering
        \includegraphics[width=\textwidth]{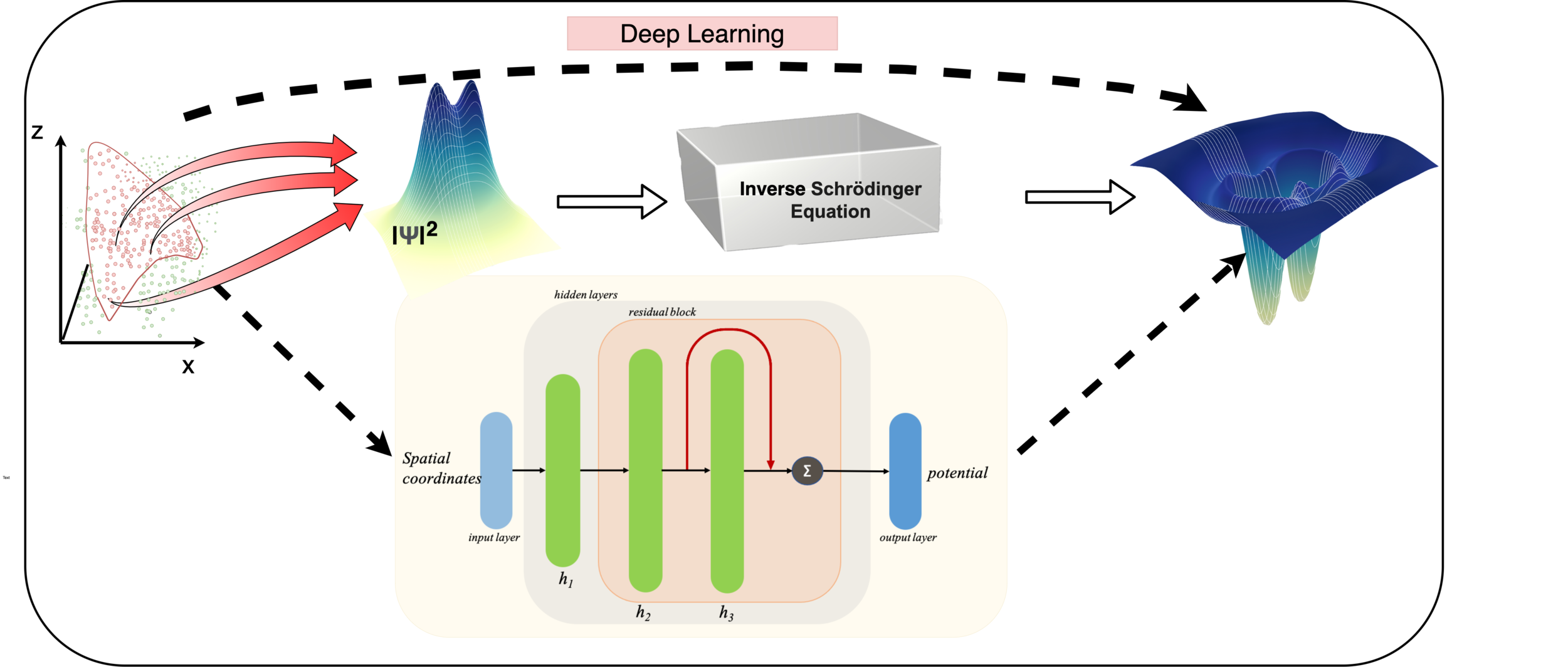} 
        \caption{Graphical representation of our framework} \label{fig:workflow}
    \end{figure*}

\subsection{Predicting potentials}
The usual method for describing systems in quantum mechanics is by obtaining the wave--function of the system as a solution of the Schr\"{o}dinger equation. These wave--functions strongly depend on the Hamiltonian, and in particular, the definition of the potential used to describe the system. However, one could also describe a quantum phenomena through the solution of the inverse problem, i.e. by finding an effective potential or function that contains all the important physical constrains that generated the observed outcomes. Inverse problems like this one are common in quantum mechanics and electronic structure, for example, the field of DFT~\cite{jensen2018numerical,parr1995density,Burke_2005} has, at its core, this type of inverse problems. 
As discussed before, solutions for the full Schr\"{o}dinger's equation, and thus wave--functions, are  difficult to obtain except for some simple  models;  the  probability density $|\psi(\Vec{x})|^2$, on the other hand, may be experimentally inferred for several quantum systems. Thus, an approximated wave--function, $|\psi|=\sqrt{|\psi(\Vec{x})|^2}$, may be defined for  the construction of the effective potential.   
%


\section{Quantum Potential Neural Networks}
In this section the proposed NN and the loss function that were used to compute the effective potentials of various quantum systems are described. This section is divided into two parts: (\textbf{i}) Time--independent systems and (\textbf{ii}) Time--dependent systems
\subsection{Time--independent Systems}
In order to obtain the effective potential, a new parametric function $U_{\theta}$ is learned in an unsupervised manner. This parametric function corresponds to the effective potential of the quantum system and  was obtained by implementing a loss function that obeys the time--independent Schr\"{o}dinger (TISE) equation (Eq.~\ref{eqn:TISE}),  
\begin{equation}\label{eqn:time_independent_Schrodinger_loss}
    L_{TISE}(\theta) = \bigg\lvert\bigg\rvert D \left(-\frac{\hbar^{2}}{2m} \frac{\Delta_{\Vec{x}} |\psi|}{|\psi|} + U_{\theta}(\Vec{x})\right)\bigg\lvert\bigg\rvert_{2}^2,
\end{equation}
 where $D$ is the total derivative operator acting on multi-variate function $-\frac{\hbar^{2}}{2m} \frac{\Delta_{\Vec{x}} |\psi|}{|\psi|} + U_{\theta}(\Vec{x})$  and $\lvert\rvert \cdot \lvert\rvert_{2}$ is the Frobenius norm. Because of the definition of  this loss function, energy conservation is effectively demanded for time--independent systems. Since the $U_{\theta}$ function is given by a differential equation, an initial condition was added to ensure that a unique function is learned. The initial  conditions used for the different quantum systems are based on the inherent nature of each of the systems, more information and explanations about some of these conditions can be found in the literature~\cite{romanowski2007numerical}. Finally, the loss function after the consideration of the initial condition reads
 \begin{equation}\label{eqn:TISE_loss}
     L(\theta) = L_{TISE}(\theta) + (U_{\theta}(\Vec{x}) - y)^2
 \end{equation}
 where $\Vec{x}$ is some point in the domain of the function and $y$ is the expected ground truth value of the true potential at that point. \\
 The main observation here is that using $|\psi|$ (instead of $\psi$) to solve for the potential leads to the correct potential except, possibly, at finitely many points where $\psi$ changes signs. However, this does not create any difficulties for training the proposed model.  
 
 \subsection{Time--dependent Systems}
For time--dependent systems, the formulation for the time--dependent Schr\"odinger equation (TDSE) loss reads 
\begin{equation}\label{eqn:TDSE_loss}
L_{TDSE}(\theta) = \bigg\vert\bigg\vert Re\bigg(\frac{i\frac{\partial \psi}{\partial t} + \frac{\hbar^{2}}{2m} \frac{\partial^2 \psi}{\partial \Vec{x}^2}}{\psi}\bigg) - U_{\theta} \bigg\vert \bigg\vert_{2}^{2}.
\end{equation}
It is important to mention that complex numbers may be more common to appear in the time--dependent solution of the Schr\"odinger equation; however, for the current study, the probability density of the considered systems are described  with an Hermitian Hamiltonian, and thus only real observables were considered to avoid the handling of complex values. For the time--dependent results presented in this report, the  QPNN was trained  with the full wave--function instead of just the probability density. In a future work, the density--to--potential results for time--dependent systems will be explored and discussed in detail. 
 
\subsection{Model Architecture} For the construction of the NN, a $4$ layer feedforward network with hidden sizes of $32, 128$ and $128$ with a residual connection between second and third layers was used. The input to the model are the $\Vec{x}$ spatial coordinates for time independent systems and ($\Vec{x}, t$) for time dependent systems. For the network training, $3000$ of these coordinates were randomly selected from the domain of definition of each particular system. The model was  trained for $500$ epochs with Adam optimizer~\cite{kingma2017adam}.

\begin{table*}[t]
\renewcommand{\arraystretch}{0.5}
\begin{center}
\caption{A quantitative analysis for the QPNN}
\label{tab:rmse}
 \resizebox{\textwidth}{!}{\begin{tabular}{|l c c c c |} 
 \hline
 System & \begin{tabular}[c]{@{}c@{}} RMSE between True and Learned \\ Potentials (QPNN) \end{tabular} & \begin{tabular}[c]{@{}c@{}} RMSE between True and Learned \\ Potentials (using RK4) \end{tabular} & \begin{tabular}[c]{@{}c@{}}  RMSE between True and Learned \\ Energies (QPNN) \end{tabular} & \begin{tabular}[c]{@{}c@{}}  RMSE between True and Learned \\ Energies (RK4) \end{tabular} \\ [0.5ex] 
 \hline\hline
 Harmonic Oscillator& $\num{1e-2} \pm \num{5e-3}$& $\num{9e-3} \pm \num{4e-4}$ &  $\num{1e-2} \pm \num{2e-3}$ & $\num{5e-2} \pm \num{7e-3}$\\[0.9ex] 
 P\"oschl--Teller potential& $\num{1e-4} \pm \num{6e-5}$& $\num{2e-4} \pm \num{3e-5}$ & $\num{8e-4} \pm \num{6e-5}$ & $\num{7e-3} \pm \num{8e-4}$ \\[0.9ex]
 
 
 $\rm H_2$ molecule & $\num{2e-3} \pm \num{4e-4}$& $\num{3e-3} \pm \num{2e-4}$ & $\num{9e-3} \pm \num{7e-4}$ & $\num{4e-3} \pm \num{2e-4}$\\[0.9ex]
 
 Soliton & $\num{3e-2} \pm \num{4e-3}$& - & -& -\\[0.9ex]

 \hline
\end{tabular}}
\end{center}
\end{table*}
\section{Related Work}
The use of deep learning for understating physical phenomena has been an active field of development.  In particular, there is a considerable amount of literature where authors have endowed neural networks with classical Hamiltonian mechanics~\cite{toth2019hamiltonian,greydanus2019hamiltonian,tong2020symplectic, Iten_2020, bondesan2019learning, zhong2019symplectic,chmiela2017machine}; conservation of energy and irreversibility in time are the key features of such networks. There are recent reports extending these results in cases of damped pendula, i.e., systems where there is dissipation of energy~\cite{zhong2020dissipative}. In computational quantum mechanics, deep neural networks have been implemented to learn representations and extract the necessary features to predict desired properties from raw unprocessed data~\cite{goh2017deep}. Recently, two methods for estimating the density matrix for a quantum system, the 
Quantum Maximum Likelihood (QML) and Quantum Variational Inference (QVI) method, were introduced~\cite{cranmer2019inferring}. For these methods, the authors  used a flow based method~\cite{toth2019hamiltonian,rezende2015variational} to increase the expressivity of their variational family of density matrices. The applicability of these methods, however, has been only validated for the harmonic and anharmonic quantum oscillator models.
Applications of deep learning to quantum mechanics is still in its early stages~\cite{torfi2020natural,raissi2017physicsI, raissi2017physicsII, raissi2019physics, dai2020machine, Carleo_2019,amabilino2019training,Unke2019MachineLP,schmitz2019machine,schmidt2017predicting, hibatallah2020recurrent,nakajima2020neural,pu2020soliton,mills2017deep,manzhos2020machine}. Most of the deep learning quantum mechanic frameworks introduced so far are focused on either solving the Schr\"odinger equation or predicting the trends of specific observables such as the system's energy. Concerning inverse problems, Raissi et al, introduced  the  physics--informed neural network for solving forward and inverse problems involving nonlinear partial differential equations~\cite{raissi2019physics}.  Although the impressive results reported in this work, in terms of  inverse problem solutions, the deep learning framework reported by Raissi et al is focused only on the solution of a partial
differential equation for the  prediction of pressure profiles in an classical system. On the other hand, in quantum mechanics observables may be inferred when a valid effective potential is known for a given quantum system; thus, the solution of the density--to--potential inversion problem to  predict effective potential functions~\cite{jensen2018numerical} play an important role in the understanding of the quantum phenomena, and in particular in the elucidation of  the electronic structure of molecules  from a density functional theory perspective.
In this regard, Nagai and coworkers have proposed the Neural--network Kohn--Sham exchange--correlation potential~\cite{nagai2018neural}, which propose a supervised training scheme that uses information from  well defined potentials and probability densities to train a NN. However, to the best of our knowledge, there are no reported works that uses deep learning to solve the density--to--potential \textit{inverse problem} to systematically estimate potentials from observations in a completely \textit{unsupervised} manner. 
\section{Experiments}
The performance of the proposed Quantum Potential Neural Network is validated on four different quantum systems, one of these systems describes the temporal evolution of a quantum wave whereas the other three are examples of time--independent systems. Among the time--independent systems, exact analytical solutions for the time--independent Schr\"odinger equation can be obtained only for the harmonic oscillator and the P\"oschl--Teller (PT) potentials whereas for the $\text{H}_2$ molecule, only approximate solutions are attained. 
Details about the solutions, physical implications and interpretations of these systems can be found in any standard book on quantum mechanics~\cite{Sakurai,robinett,Pronchik}. Finally, the potentials learned by the QPNN were used to compute the total energy of each of the systems. The quantitative results for all the reported experiments are summarized in table \ref{tab:rmse}. In order to compare the results obtained (in the form of solutions to a differential equation) via the NN techniques to those obtained through well established approaches, the differential equations were solved numerically using the well--known and standard Runge--Kutta 4th Order (or RK4) integrator implemented in the standard python libraries. The RK4 algorithm provides means of solving various kinds of differential equations and is generally considered as a robust workhorse to bench mark new computational techniques~\cite{LandauPaez}. The differences in accuracy of the values obtained by both the proposed QPNN method and the RK4 numerical integrator were quantified through their root mean square error (RMSE) values.
 \subsection{Density--to--Potential Experiments}
For  the density--to--potential experiments, the exact wave--functions, $\psi(\Vec{x})$, for the quantum Harmonic oscillator and the PT potential were obtained by solving the time--independent Schr\"odinger equation. These wave--functions were later used to define the probability distribution, $|\psi(\Vec{x})|^2$, for each of the systems. In the case of the Hydrogen molecule, the probability density was defined according to the one--electron $1s$ orbital function that delineate the approximated electronic density for the Hydrogen molecule in a Born--Oppenheimer approximation.  These probability densities were used to define, for each of the quantum systems, an approximated probability amplitude function,  $|\psi|=\sqrt{|\psi(\Vec{x})|^2}$. The effective potential function for each of the systems was obtained  by training the QPNN  with information provided by randomly selected coordinates evaluated onto the  approximated probability amplitudes and into the loss function.
\subsubsection*{Quantum Harmonic Oscillator (QHO):} 
\begin{figure}[h!]
    \centering
        \includegraphics[width=.85\columnwidth]{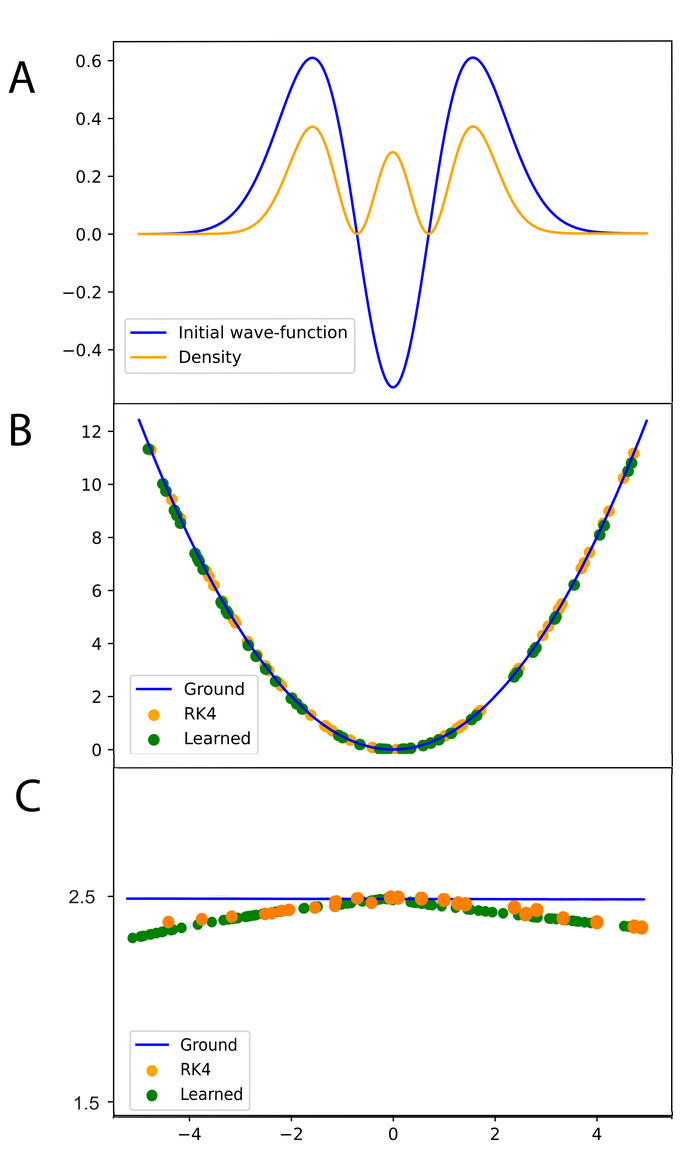} 
      \caption{Harmonic Oscillator system. (A) Wave--function and probability Density used in the QPNN, (B) Ground and Learned Potential, (C) Total Energy} \label{fig:pot_dist}
    \end{figure}

The motion of the the one--dimension QHO is, perhaps, the simplest quantum mechanical system whose motion follows a linear differential equation with constant coefficients.
In the QPNN framework, for the prediction of the QHO potential, the coordinate variable $\Vec{x}$, was randomly sampled from $[-5,5]$, and was used as the input to the model. In the analytical solution of the time--independent Schr\"odinger equation, the wave--functions for the different states of the  QHO are given by Hermite polynomials $H_n, n=0,1,\cdots$, whereas the energies corresponding to these states depends on the force constant $w$ and are given by $E_{n}=\hbar w (n+1/2)$. The analytical wave--functions for the different states of the QHO defined the probability densities used to train the QPNN.  
In this case, the initial condition imposed is the fact that at the zero point of the reference coordinates, all the energy in the system is kinetic, and thus the potential energy at this point is zero, i.e. $V(0)=0$. Fig~\ref{fig:pot_dist} shows the used probability density, the learned potential and the energy computed by the QPNN. 
    \subsubsection*{P\"oschl-Teller potential:} 
\begin{figure}[h!]
    \centering
        \includegraphics[width=.85\columnwidth]{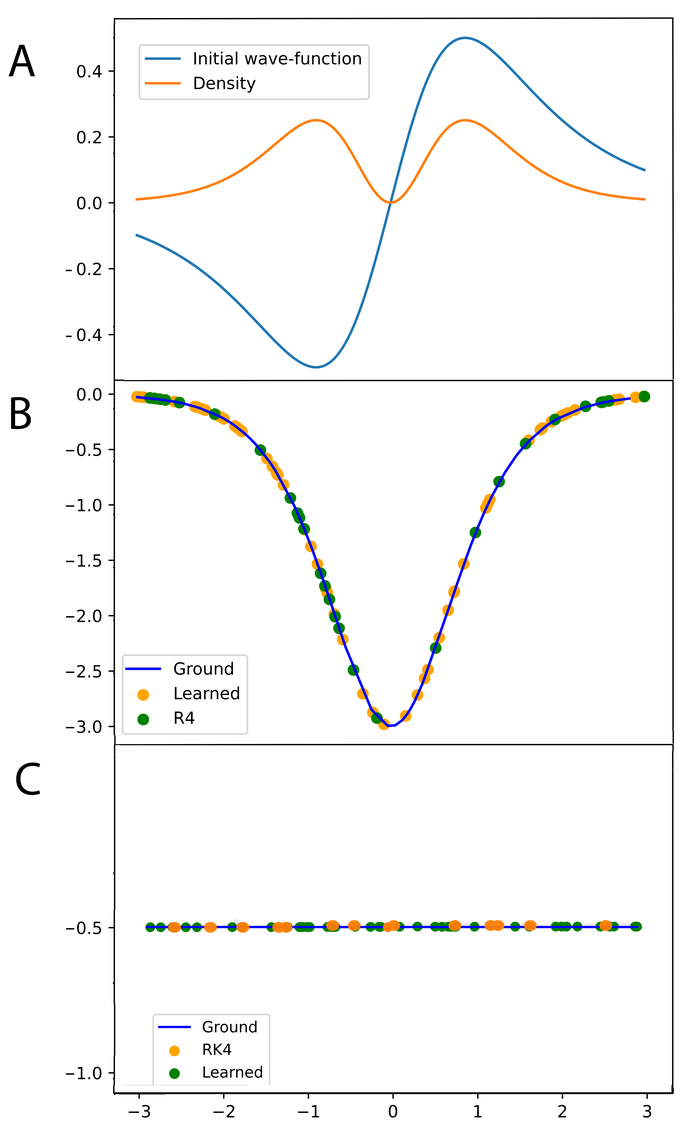}

    \caption{P\"oschl Teller system. (A) Wave--function and Density, (B) Ground and Learned Potential, (C) Total Energy} \label{fig:pot_poschl}
    \end{figure}

The PT potential
is a special class of anharmonic potentials for which the one--dimensional Schr\"odinger equation can be solved in terms of special functions. 
This potential may be used to model  vibrational molecular potentials with out--of--plane bending vibrations~\cite{Penn,JiaZhangPeng} and observables of diatomic potentials~\cite{Pronchik}. 
For the PT potential,  the wave--functions used to define  the approximated probability amplitude function are the Legendre functions  $P^{\mu}_{\lambda}$ \cite{riley_1974} with energy eigenvalues $E_{\mu}$  and potential depth $V_0$\cite{BrowndelaPena},
\begin{equation}
\Big\{P^{\mu}_{\lambda}(\tanh x) \;\Big|\; E_{\mu}=\frac{-\mu^{2}}{2}, V_0=\frac{-\lambda(\lambda+1)}{2},\; 
\substack{\lambda = 1,2, \cdots \\
\;\mu = 1,2, \cdots, \lambda
} \Big\}.
\end{equation}
Details about several suitable boundary terms and initial conditions for this type of potentials are formulated and reviewed in the literature~\cite{Agboola,BrowndelaPena}. The input for this experiment is the spatial coordinate $x$ randomly sampled from $[-3,3]$. For this experiment, the wave--function defined by the Legendre function with $\lambda = 2$ and $\mu = 1$ was employed. 
From the density, the initial wave--function takes the form; 
\begin{equation}\psi^1_2\left(x \right) = |\text{tanh } x| \text{ sech }x
\end{equation}
 Fig~\ref{fig:pot_poschl} shows the probability density used to train the system, as well as the learned potential and energy computed for the system. 
    
\begin{figure*}[t]
    \centering
        \includegraphics[width=\textwidth]{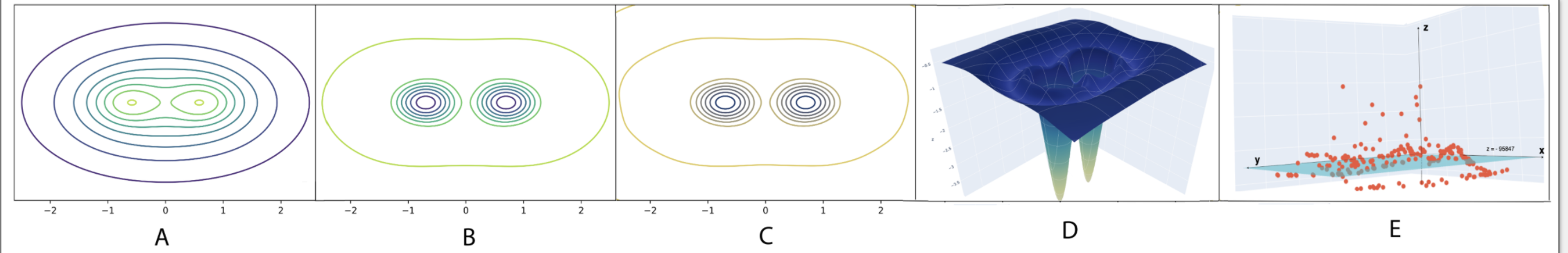} 
        \caption{$\text{H}_2$ molecule system. (A) Density, (B) Ground Truth Potential, (C,D) Learned Potential, (E) Computed Energy} \label{fig:manybody}
    \end{figure*}
\subsubsection*{Hydrogen molecule:} The  $\rm H_2$  molecule is the first multi-electronic system with approximated probability density considered. For the training of the QPNN, an ab initio electron density for the $\rm H_2$ molecule with an equilibrium bond length ($x_{r_e}$) of 1.346{\AA} and total energy of -0.958470046928 a.u. was approximated by using a fast and systematic self--consistent field method~\cite{helgaker2014molecular}. This density was computed using three Gaussian primitive functions for each $\rm H$ atom where the $\Vec{x}$ coordinate defined on $[-3,3]$ was chosen as the reference internal coordinate. The initial conditions for the system were defined following the same lines as in~\cite{rafi1995empirical}; specifically, the fact that $V\left(x_{r_e}\right)=0$.  Fig~\ref{fig:manybody} shows the probability density used to train the QPNN, as well as the learned potential and the energy computed using this potential.  

\subsection{Exploration of a Time--Dependent System}
\begin{figure}[h!]
    \centering
        \includegraphics[width=\columnwidth]{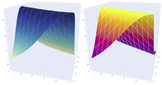} 
      \caption{Left: Soliton Ground Truth Potential, Right: Soliton Learned Potential} \label{fig:3d_figs}
    \end{figure}

In order to explore the behavior of the QPNN in systems with dependence on time, the effective potential for a soliton model was computed. Solitons represent solitary waves propagating without any temporal evolution in shape or
size when viewed in the reference frame moving with the group velocity of the waves~\cite{PDE_soliton}. 
This type of solitary waves are  particularly important in the Bose--Einstein condensation theory and arise in many contexts such as  the elevation of the surface of water and the intensity of light in optical fibers. Solitons form a special class of solutions of model equations, including the Korteweg de--Vries (KdV) and the Nonlinear Schr\"odinger (NLS) equations. In this particular experiment, the one--dimensional
soliton satisfies the following differential equation: 
\begin{equation}
    i\frac{\partial \psi}{\partial t} + \frac{\partial^2 \psi}{\partial x^2} + U(x,t)\psi = 0.
\end{equation}
Thus, for this system, the loss function used to train the QPNN is given by equation~\ref{eqn:TDSE_loss} where $\psi = 2\text{sech}(\sqrt{2}(x-2t))e^{i(x+t)}$ and $U(x,t)$ is $|\psi|^2$. For this experiment, the coordinates for the QPNN input were defined on  $\Vec{x},\Vec{t} \in[0,1]$. Fig~\ref{fig:3d_figs} shows the potential obtained by the QPNN for this system.
\section{Discussion} 
Table~\ref{tab:rmse} reports the RMSE values for the different models studied in this work. In the case of the time--independent models, for the QPNN, the harmonic oscillator
presents the largest RMSE value between the true and learned potential whereas the PT potential has the lowest RMSE among all the time--independent systems. 
For all the systems but the harmoninc oscillator, the RMSE values for the learned potentials are comparable in magnitude with those obtained with the RK4 method. In the case of the energies, when compared against the exact energy, the RMSE values for both the QPNN and the RK4 method are around the same magnitude for all systems but the PT potential, where the RMSE for the QPNN is one order of magnitude lower than for the RK4 method. In terms of the energy values computed for each of the systems, when referenced to the exact energy (blue line) a similar trend can be observed for the energies computed with the RK4 method and the QPNN. If the RK4 method is regarded as a more robust and mathematically superior method for the calculation of the effective potential function, this trend may be interpreted as an indicator of the reliability of the QPNN. In the case of the soliton model, although  solutions using the RK4 were not feasible due to the nature of the system, the magnitude of RMSE values between the true potential and the learned potential suggests the same qualitative behaviour as the one obtained with the time--independent models.

\section{Conclusion}
In this work, the QPNN for learning the effective potential functions of different quantum systems was presented. This new neural network is capable of learning the effective potential functions of a variety of systems in a completely unsupervised manner. The results obtained for the different studied systems suggest that the QPNN has an accuracy comparable to the RK4 integrator. The potentials learned with this new QPNN can be used to calculate observables like the energy of the system. The use of the QPNN was  also extended  for time--dependent systems where the full wave--function was used. In the time--dependent case, $|\psi|$ can not simply be taken as a proxy for a wave--function as one needs to take into account some phase information, i.e., the wave--function in that case can not be a real valued function. In future work, the  density--to--potential problem will be analysed by incorporating  phase information to create a suitable proxy wave--function.

\bibliography{ref}

\begin{thebibliography}{70}
\providecommand{\natexlab}[1]{#1}
\providecommand{\url}[1]{\texttt{#1}}
\providecommand{\urlprefix}{URL }
\expandafter\ifx\csname urlstyle\endcsname\relax
  \providecommand{\doi}[1]{doi:\discretionary{}{}{}#1}\else
  \providecommand{\doi}{doi:\discretionary{}{}{}\begingroup
  \urlstyle{rm}\Url}\fi

\bibitem[{Agboola(2010)}]{Agboola}
Agboola, D. 2010.
\newblock Solutions to the Modified Pöschl{\textendash}Teller Potential in
  D-Dimensions.
\newblock \emph{Chinese Physics Letters} 27(4): 040301.

\bibitem[{Amabilino et~al.(2019)Amabilino, Bratholm, Bennie, Vaucher, Reiher,
  and Glowacki}]{amabilino2019training}
Amabilino, S.; Bratholm, L.~A.; Bennie, S.~J.; Vaucher, A.~C.; Reiher, M.; and
  Glowacki, D.~R. 2019.
\newblock Training neural nets to learn reactive potential energy surfaces
  using interactive quantum chemistry in virtual reality.
\newblock \emph{The Journal of Physical Chemistry A} 123(20): 4486--4499.

\bibitem[{Aster, Borchers, and Thurber(2018)}]{aster2018parameter}
Aster, R.~C.; Borchers, B.; and Thurber, C.~H. 2018.
\newblock \emph{Parameter estimation and inverse problems}.
\newblock Elsevier.

\bibitem[{Athanassoulis(2008)}]{athanassoulis2008exact}
Athanassoulis, A.~G. 2008.
\newblock Exact equations for smoothed Wigner transforms and homogenization of
  wave propagation.
\newblock \emph{Applied and Computational Harmonic Analysis} 24(3): 378--392.

\bibitem[{Beals and Greiner(2009)}]{beals2009strings}
Beals, R.; and Greiner, P.~C. 2009.
\newblock Strings, waves, drums: spectra and inverse problems.
\newblock \emph{Analysis and Applications} 7(02): 131--183.

\bibitem[{Bondesan and Lamacraft(2019)}]{bondesan2019learning}
Bondesan, R.; and Lamacraft, A. 2019.
\newblock Learning Symmetries of Classical Integrable Systems.

\bibitem[{Burke, Werschnik, and Gross(2005)}]{Burke_2005}
Burke, K.; Werschnik, J.; and Gross, E. K.~U. 2005.
\newblock Time-dependent density functional theory: Past, present, and future.
\newblock \emph{The Journal of Chemical Physics} 123(6): 062206.
\newblock ISSN 1089-7690.
\newblock \doi{10.1063/1.1904586}.
\newblock \urlprefix\url{http://dx.doi.org/10.1063/1.1904586}.

\bibitem[{Carleo et~al.(2019)Carleo, Cirac, Cranmer, Daudet, Schuld, Tishby,
  Vogt-Maranto, and Zdeborová}]{Carleo_2019}
Carleo, G.; Cirac, I.; Cranmer, K.; Daudet, L.; Schuld, M.; Tishby, N.;
  Vogt-Maranto, L.; and Zdeborová, L. 2019.
\newblock Machine learning and the physical sciences.
\newblock \emph{Reviews of Modern Physics} 91(4).
\newblock ISSN 1539-0756.
\newblock \doi{10.1103/revmodphys.91.045002}.
\newblock \urlprefix\url{http://dx.doi.org/10.1103/RevModPhys.91.045002}.

\bibitem[{Case(2008{\natexlab{a}})}]{case_Wigner}
Case, W.~B. 2008{\natexlab{a}}.
\newblock Wigner functions and Weyl transforms for pedestrians.
\newblock \emph{American Journal of Physics} 76(10): 937--946.
\newblock \doi{10.1119/1.2957889}.
\newblock \urlprefix\url{https://doi.org/10.1119/1.2957889}.

\bibitem[{Case(2008{\natexlab{b}})}]{case2008wigner}
Case, W.~B. 2008{\natexlab{b}}.
\newblock Wigner functions and Weyl transforms for pedestrians.
\newblock \emph{American Journal of Physics} 76(10): 937--946.

\bibitem[{Chadan and Sabatier(2012)}]{chadan2012inverse}
Chadan, K.; and Sabatier, P.~C. 2012.
\newblock \emph{Inverse problems in quantum scattering theory}.
\newblock Springer Science \& Business Media.

\bibitem[{Chen, Xiong, and Shao(2018)}]{Chen_2018}
Chen, Z.; Xiong, Y.; and Shao, S. 2018.
\newblock Numerical Methods for the Wigner Equation with Unbounded Potential.
\newblock \emph{Journal of Scientific Computing} 79(1): 345–368.
\newblock ISSN 1573-7691.
\newblock \doi{10.1007/s10915-018-0853-0}.
\newblock \urlprefix\url{http://dx.doi.org/10.1007/s10915-018-0853-0}.

\bibitem[{Chen, Xiong, and Shao(2019)}]{chen2019numerical}
Chen, Z.; Xiong, Y.; and Shao, S. 2019.
\newblock Numerical methods for the Wigner equation with unbounded potential.
\newblock \emph{Journal of Scientific Computing} 79(1): 345--368.

\bibitem[{Chmiela et~al.(2017)Chmiela, Tkatchenko, Sauceda, Poltavsky,
  Sch{\"u}tt, and M{\"u}ller}]{chmiela2017machine}
Chmiela, S.; Tkatchenko, A.; Sauceda, H.~E.; Poltavsky, I.; Sch{\"u}tt, K.~T.;
  and M{\"u}ller, K.-R. 2017.
\newblock Machine learning of accurate energy-conserving molecular force
  fields.
\newblock \emph{Science advances} 3(5): e1603015.

\bibitem[{Cranmer, Golkar, and Pappadopulo(2019)}]{cranmer2019inferring}
Cranmer, K.; Golkar, S.; and Pappadopulo, D. 2019.
\newblock Inferring the quantum density matrix with machine learning.

\bibitem[{Cranmer et~al.(2020)Cranmer, Greydanus, Hoyer, Battaglia, Spergel,
  and Ho}]{cranmer2020lagrangian}
Cranmer, M.; Greydanus, S.; Hoyer, S.; Battaglia, P.; Spergel, D.; and Ho, S.
  2020.
\newblock Lagrangian Neural Networks.

\bibitem[{Curtright, Fairlie, and Zachos(1998)}]{curtright1998features}
Curtright, T.; Fairlie, D.; and Zachos, C. 1998.
\newblock Features of time-independent Wigner functions.
\newblock \emph{Physical Review D} 58(2): 025002.

\bibitem[{Dai et~al.(2020)Dai, Jasinski, Montaner, Forrey, Yang, Stancil,
  Balakrishnan, Vargas-Hern{\'a}ndez, and Krems}]{dai2020machine}
Dai, J.; Jasinski, A.; Montaner, J.; Forrey, R.; Yang, B.; Stancil, P.;
  Balakrishnan, N.; Vargas-Hern{\'a}ndez, R.; and Krems, R. 2020.
\newblock Machine-learning-corrected quantum dynamics calculations.
\newblock \emph{Bulletin of the American Physical Society} .

\bibitem[{Du and Narasimhan(2019)}]{du2019taskagnostic}
Du, Y.; and Narasimhan, K. 2019.
\newblock Task-Agnostic Dynamics Priors for Deep Reinforcement Learning.

\bibitem[{Feynman, Leighton, and Sands(1965)}]{feynman1965feynman}
Feynman, R.; Leighton, R.; and Sands, M. 1965.
\newblock The Feynman Lectures on Physics Vol. III, chap. 21, sec. 21-9.

\bibitem[{Galleani and Cohen(2002)}]{galleani2002approximation}
Galleani, L.; and Cohen, L. 2002.
\newblock Approximation of the Wigner distribution for dynamical systems
  governed by differential equations.
\newblock \emph{EURASIP Journal on Advances in Signal Processing} 2002(1):
  514609.

\bibitem[{Goh, Hodas, and Vishnu(2017)}]{goh2017deep}
Goh, G.~B.; Hodas, N.~O.; and Vishnu, A. 2017.
\newblock Deep learning for computational chemistry.
\newblock \emph{Journal of computational chemistry} 38(16): 1291--1307.

\bibitem[{Gomes and Silva(2008)}]{gomes2008wigner}
Gomes, D.~A.; and Silva, J.~D. 2008.
\newblock On the Wigner transform of solutions to the Schrodinger equation.
\newblock \emph{S{\~a}o Paulo Journal of Mathematical Sciences} 2(1): 85--97.

\bibitem[{Greydanus, Dzamba, and Yosinski(2019)}]{greydanus2019hamiltonian}
Greydanus, S.; Dzamba, M.; and Yosinski, J. 2019.
\newblock Hamiltonian neural networks.
\newblock In \emph{Advances in Neural Information Processing Systems},
  15353--15363.

\bibitem[{Groetsch and Groetsch(1993)}]{groetsch1993inverse}
Groetsch, C.~W.; and Groetsch, C. 1993.
\newblock \emph{Inverse problems in the mathematical sciences}, volume~52.
\newblock Springer.

\bibitem[{Helgaker, Jorgensen, and Olsen(2014)}]{helgaker2014molecular}
Helgaker, T.; Jorgensen, P.; and Olsen, J. 2014.
\newblock \emph{Molecular electronic-structure theory}.
\newblock John Wiley \& Sons.

\bibitem[{Heller(1976)}]{heller1976wigner}
Heller, E.~J. 1976.
\newblock Wigner phase space method: Analysis for semiclassical applications.
\newblock \emph{The Journal of Chemical Physics} 65(4): 1289--1298.

\bibitem[{Hernández de~la Peña(2018)}]{BrowndelaPena}
Hernández de~la Peña, L. 2018.
\newblock A Simplified Pöschl–Teller Potential: An Instructive Exercise for
  Introductory Quantum Mechanics.
\newblock \emph{Journal of Chemical Education} 95(11): 1989--1995.
\newblock \doi{10.1021/acs.jchemed.8b00029}.

\bibitem[{Hibat-Allah et~al.(2020)Hibat-Allah, Ganahl, Hayward, Melko, and
  Carrasquilla}]{hibatallah2020recurrent}
Hibat-Allah, M.; Ganahl, M.; Hayward, L.~E.; Melko, R.~G.; and Carrasquilla, J.
  2020.
\newblock Recurrent Neural Network Wavefunctions.

\bibitem[{Higgins et~al.(2016)Higgins, Matthey, Glorot, Pal, Uria, Blundell,
  Mohamed, and Lerchner}]{higgins2016early}
Higgins, I.; Matthey, L.; Glorot, X.; Pal, A.; Uria, B.; Blundell, C.; Mohamed,
  S.; and Lerchner, A. 2016.
\newblock Early Visual Concept Learning with Unsupervised Deep Learning.

\bibitem[{Iten et~al.(2020)Iten, Metger, Wilming, del Rio, and
  Renner}]{Iten_2020}
Iten, R.; Metger, T.; Wilming, H.; del Rio, L.; and Renner, R. 2020.
\newblock Discovering Physical Concepts with Neural Networks.
\newblock \emph{Physical Review Letters} 124(1).
\newblock ISSN 1079-7114.
\newblock \doi{10.1103/physrevlett.124.010508}.
\newblock \urlprefix\url{http://dx.doi.org/10.1103/PhysRevLett.124.010508}.

\bibitem[{Jensen and Wasserman(2018)}]{jensen2018numerical}
Jensen, D.~S.; and Wasserman, A. 2018.
\newblock Numerical methods for the inverse problem of density functional
  theory.
\newblock \emph{International Journal of Quantum Chemistry} 118(1): e25425.

\bibitem[{Jia, Zhang, and Peng(2017)}]{JiaZhangPeng}
Jia, C.-S.; Zhang, L.-H.; and Peng, X.-L. 2017.
\newblock Improved Pöschl–Teller potential energy model for diatomic
  molecules.
\newblock \emph{International Journal of Quantum Chemistry} 117(14): e25383.
\newblock \doi{https://doi.org/10.1002/qua.25383}.

\bibitem[{Kingma and Ba(2017)}]{kingma2017adam}
Kingma, D.~P.; and Ba, J. 2017.
\newblock Adam: A Method for Stochastic Optimization.

\bibitem[{Klimov, Sainz, and Romero(2020)}]{klimov2020truncated}
Klimov, A.; Sainz, I.; and Romero, J. 2020.
\newblock Truncated Wigner approximation as non-positive Kraus map.
\newblock \emph{Physica Scripta} 95(7): 074006.

\bibitem[{Kolesnikov et~al.(2019)Kolesnikov, Beyer, Zhai, Puigcerver, Yung,
  Gelly, and Houlsby}]{alex2019big}
Kolesnikov, A.; Beyer, L.; Zhai, X.; Puigcerver, J.; Yung, J.; Gelly, S.; and
  Houlsby, N. 2019.
\newblock Big Transfer (BiT): General Visual Representation Learning.

\bibitem[{Lample and Charton(2019)}]{lample2019deep}
Lample, G.; and Charton, F. 2019.
\newblock Deep Learning for Symbolic Mathematics.

\bibitem[{Landau, Paez, and Bordeianu(2015)}]{LandauPaez}
Landau, R.; Paez, M.~J.; and Bordeianu, C. 2015.
\newblock \emph{Computational Physics: Problem Solving with Python}.
\newblock Wiley, 3rd edition.

\bibitem[{Manzhos(2020)}]{manzhos2020machine}
Manzhos, S. 2020.
\newblock Machine learning for the solution of the Schr{\"o}dinger equation.
\newblock \emph{Machine Learning: Science and Technology} 1(1): 013002.

\bibitem[{Mills, Spanner, and Tamblyn(2017)}]{mills2017deep}
Mills, K.; Spanner, M.; and Tamblyn, I. 2017.
\newblock Deep learning and the Schr{\"o}dinger equation.
\newblock \emph{Physical Review A} 96(4): 042113.

\bibitem[{Nagai et~al.(2018)Nagai, Akashi, Sasaki, and
  Tsuneyuki}]{nagai2018neural}
Nagai, R.; Akashi, R.; Sasaki, S.; and Tsuneyuki, S. 2018.
\newblock Neural-network Kohn-Sham exchange-correlation potential and its
  out-of-training transferability.
\newblock \emph{The Journal of chemical physics} 148(24): 241737.

\bibitem[{Nakajima, Tanaka, and Hashimoto(2020)}]{nakajima2020neural}
Nakajima, M.; Tanaka, K.; and Hashimoto, T. 2020.
\newblock Neural Schr\"{o}dinger Equation: Physical Law as Neural Network.
\newblock \emph{arXiv preprint arXiv:2006.13541} .

\bibitem[{Nakatsuji(2002)}]{nakatsuji2002inverse}
Nakatsuji, H. 2002.
\newblock Inverse Schr{\"o}dinger equation and the exact wave function.
\newblock \emph{Physical Review A} 65(5): 052122.

\bibitem[{Parr and Yang(1995)}]{parr1995density}
Parr, R.~G.; and Yang, W. 1995.
\newblock Density-functional theory of the electronic structure of molecules.
\newblock \emph{Annual Review of Physical Chemistry} 46(1): 701--728.

\bibitem[{Pronchik and Williams(2003)}]{Pronchik}
Pronchik, J.~N.; and Williams, B.~W. 2003.
\newblock Exactly Solvable Quantum Mechanical Potentials: An Alternative
  Approach.
\newblock \emph{Journal of Chemical Education} 80(8): 918.
\newblock \doi{10.1021/ed080p918}.

\bibitem[{Pu, Li, and Chen(2020)}]{pu2020soliton}
Pu, J.; Li, J.; and Chen, Y. 2020.
\newblock Soliton, Breather and Rogue Wave Solutions for Solving the Nonlinear
  Schr\"odinger Equation Using a Deep Learning Method with Physical
  Constraints.
\newblock \emph{arXiv preprint arXiv:2011.04949} .

\bibitem[{Rafi et~al.(1995)}]{rafi1995empirical}
Rafi, M.; et~al. 1995.
\newblock An empirical potential function of diatomic molecules.
\newblock \emph{Physics Letters A} 205(5-6): 383--387.

\bibitem[{Raissi, Perdikaris, and
  Karniadakis(2017{\natexlab{a}})}]{raissi2017physicsI}
Raissi, M.; Perdikaris, P.; and Karniadakis, G.~E. 2017{\natexlab{a}}.
\newblock Physics Informed Deep Learning (Part I): Data-driven Solutions of
  Nonlinear Partial Differential Equations.
\newblock \emph{arXiv preprint arXiv:1711.10561} .

\bibitem[{Raissi, Perdikaris, and
  Karniadakis(2017{\natexlab{b}})}]{raissi2017physicsII}
Raissi, M.; Perdikaris, P.; and Karniadakis, G.~E. 2017{\natexlab{b}}.
\newblock Physics Informed Deep Learning (Part II): Data-driven Discovery of
  Nonlinear Partial Differential Equations.
\newblock \emph{arXiv preprint arXiv:1711.10566} .

\bibitem[{Raissi, Perdikaris, and Karniadakis(2019)}]{raissi2019physics}
Raissi, M.; Perdikaris, P.; and Karniadakis, G.~E. 2019.
\newblock Physics-informed neural networks: A deep learning framework for
  solving forward and inverse problems involving nonlinear partial differential
  equations.
\newblock \emph{Journal of Computational Physics} 378: 686--707.

\bibitem[{Rezende and Mohamed(2015)}]{rezende2015variational}
Rezende, D.~J.; and Mohamed, S. 2015.
\newblock Variational Inference with Normalizing Flows.

\bibitem[{Riley(1974)}]{riley_1974}
Riley, K.~F. 1974.
\newblock \emph{Mathematical Methods for the Physical Sciences: An Informal
  Treatment for Students of Physics and Engineering}.
\newblock Cambridge University Press.
\newblock \doi{10.1017/CBO9781139167550}.

\bibitem[{Robinett and Robinett(2006)}]{robinett2006}
Robinett, R.; and Robinett, R.~W. 2006.
\newblock \emph{Quantum mechanics: Classical results, modern systems, and
  visualized examples}.
\newblock Oxford University Press.

\bibitem[{Robinett(1997)}]{robinett}
Robinett, R.~W. 1997.
\newblock \emph{Quantum mechanics}.
\newblock Oxford University Press, New York.

\bibitem[{Romanowski(2007)}]{romanowski2007numerical}
Romanowski, Z. 2007.
\newblock Numerical Solution Of Kohn-Sham Equation For Atom.
\newblock \emph{Acta Physica Polonica B} 38(10).

\bibitem[{Sakurai and Commins(1995)}]{Sakurai}
Sakurai, J.~J.; and Commins, E.~D. 1995.
\newblock Modern Quantum Mechanics, Revised Edition.
\newblock \emph{AAPT} .

\bibitem[{Schleder et~al.(2019)Schleder, Padilha, Acosta, Costa, and
  Fazzio}]{schleder2019dft}
Schleder, G.~R.; Padilha, A.~C.; Acosta, C.~M.; Costa, M.; and Fazzio, A. 2019.
\newblock From DFT to machine learning: recent approaches to materials
  science--a review.
\newblock \emph{Journal of Physics: Materials} 2(3): 032001.

\bibitem[{Schmidt et~al.(2017)Schmidt, Shi, Borlido, Chen, Botti, and
  Marques}]{schmidt2017predicting}
Schmidt, J.; Shi, J.; Borlido, P.; Chen, L.; Botti, S.; and Marques, M.~A.
  2017.
\newblock Predicting the thermodynamic stability of solids combining density
  functional theory and machine learning.
\newblock \emph{Chemistry of Materials} 29(12): 5090--5103.

\bibitem[{Schmitz, Godtliebsen, and Christiansen(2019)}]{schmitz2019machine}
Schmitz, G.; Godtliebsen, I.~H.; and Christiansen, O. 2019.
\newblock Machine learning for potential energy surfaces: An extensive database
  and assessment of methods.
\newblock \emph{The Journal of chemical physics} 150(24): 244113.

\bibitem[{Senn(1986)}]{Penn}
Senn, P. 1986.
\newblock The modified Poschl-Teller Oscillator.
\newblock \emph{Journal of Chemical Education} 63(1): 75.

\bibitem[{Sun et~al.(2019)Sun, Myers, Vondrick, Murphy, and
  Schmid}]{sun2019videobert}
Sun, C.; Myers, A.; Vondrick, C.; Murphy, K.; and Schmid, C. 2019.
\newblock VideoBERT: A Joint Model for Video and Language Representation
  Learning.

\bibitem[{Tong et~al.(2020)Tong, Xiong, He, Pan, and Zhu}]{tong2020symplectic}
Tong, Y.; Xiong, S.; He, X.; Pan, G.; and Zhu, B. 2020.
\newblock Symplectic Neural Networks in Taylor Series Form for Hamiltonian
  Systems.

\bibitem[{Torfi et~al.(2020)Torfi, Shirvani, Keneshloo, Tavvaf, and
  Fox}]{torfi2020natural}
Torfi, A.; Shirvani, R.~A.; Keneshloo, Y.; Tavvaf, N.; and Fox, E.~A. 2020.
\newblock Natural Language Processing Advancements By Deep Learning: A Survey.

\bibitem[{Toth et~al.(2019)Toth, Rezende, Jaegle, Racanière, Botev, and
  Higgins}]{toth2019hamiltonian}
Toth, P.; Rezende, D.~J.; Jaegle, A.; Racanière, S.; Botev, A.; and Higgins,
  I. 2019.
\newblock Hamiltonian Generative Networks.

\bibitem[{Unke and Meuwly(2019)}]{Unke2019MachineLP}
Unke, O.~T.; and Meuwly, M. 2019.
\newblock Machine Learning Potential Energy Surfaces.
\newblock \emph{arXiv: Chemical Physics} .

\bibitem[{Vogel(2002)}]{vogel2002computational}
Vogel, C.~R. 2002.
\newblock \emph{Computational methods for inverse problems}, volume~23.
\newblock Siam.

\bibitem[{Wazwaz(2009)}]{PDE_soliton}
Wazwaz, A. 2009.
\newblock \emph{Partial Differential Equations and Solitary Waves Theory}.
\newblock Springer Berlin Heidelberg.
\newblock \doi{10.1007/978-3-642-00251-9}.

\bibitem[{Zakhariev and Suzko(2012)}]{zakhariev2012direct}
Zakhariev, B.~N.; and Suzko, A.~A. 2012.
\newblock \emph{Direct and inverse problems: potentials in quantum scattering}.
\newblock Springer Science \& Business Media.

\bibitem[{Zhong, Dey, and Chakraborty(2019)}]{zhong2019symplectic}
Zhong, Y.~D.; Dey, B.; and Chakraborty, A. 2019.
\newblock Symplectic {ODE}-{N}et: Learning Hamiltonian Dynamics with Control.

\bibitem[{Zhong, Dey, and Chakraborty(2020)}]{zhong2020dissipative}
Zhong, Y.~D.; Dey, B.; and Chakraborty, A. 2020.
\newblock Dissipative Sym{ODEN}: Encoding Hamiltonian Dynamics with Dissipation
  and Control into Deep Learning.

\end{thebibliography}
\appendix

\section{Additional experiments using the full wave--function}
In order to further explore the capabilities and accuracy of our NN, we start off by describing additional experiments using the QPNN and  full wave--function as well as Wigner's functions. The modifications to our loss function is described in detail below.
\begin{table*}[t]

\begin{center}
\caption{Wave--functions, Energies, and Potentials of various time independent systems}
\label{tab:example_wave_function}
 \resizebox{\textwidth}{!}{\begin{tabular}{|l c c r|} 
 \hline
 System & Potential $V(x)$& wave--function $\psi(x)$& Energy \\ [0.5ex] 
 \hline\hline
 Harmonic Oscillator& $\frac{1}{2} kx^{2}$&${\frac {1}{\sqrt {2^{n}\,n!}}}\left({\frac {m\omega }{\pi \hbar }}\right)^{1/4} e^{-{\frac {m\omega x^{2}}{2\hbar }}} H_{n}\left({\sqrt {\frac {m\omega }{\hbar }}}x\right) $& $\hbar \omega(n+\frac{1}{2})$ \\[0.9ex] 
 P\"ochl--Teller potential& $-\frac{\lambda(\lambda +1)}{2}\text{sech}^2(x)$& $P_{\lambda}^{\mu}(\text{tanh}(x))$  & $-\frac{\hbar^{2}}{2m}(\lambda-\mu)$\\[0.9ex]
Radial Hydrogen atom& $\frac{l(l+1)}{2r^{2}}-\frac{1}{r}$ &    $e^{-r/n} (\frac{2r}{na_0})^{l}L^{2l+1}_{n+l} (\frac{2r}{na_0}) $&$-\frac{R_H}{(n+l)^2}$\\[0.9ex]
 2D Harmonic Oscillator& $\frac{1}{2} k(x^{2}
 +y^{2})$&$H_{n_{x}}\left({\sqrt {\frac {m\omega }{\hbar }}}x\right) H_{n_{y}}\left({\sqrt {\frac {m\omega }{\hbar }}}y\right)e^{-{\frac {m\omega (x^{2}+y^{2})}{2\hbar }}}$& $\hbar \omega(n_x+n_y+1)$ \\[0.9ex]
 \hline

\end{tabular}}
\end{center}
\end{table*}

\subsection{Using the 1D time--independent Schr\"odinger equation}
\begin{figure}
     \centering
        \includegraphics[width= \linewidth]{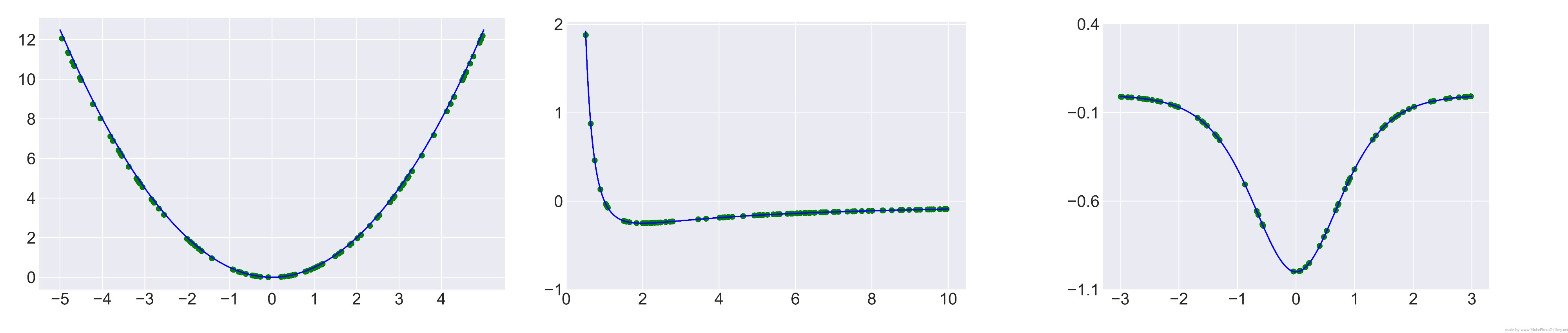} 
        \caption{Position (x-axis) vs Potential (y-axis); Ground (Blue) and Learned (Green Dots) Potentials: From left to right: Harmonic Oscillator, Hydrogen Atom (2p case), P\"oschl-Teller (1,1 case)} \label{fig:TDSE}.
    \end{figure}

\begin{figure}
     \centering
        \includegraphics[width= \linewidth]{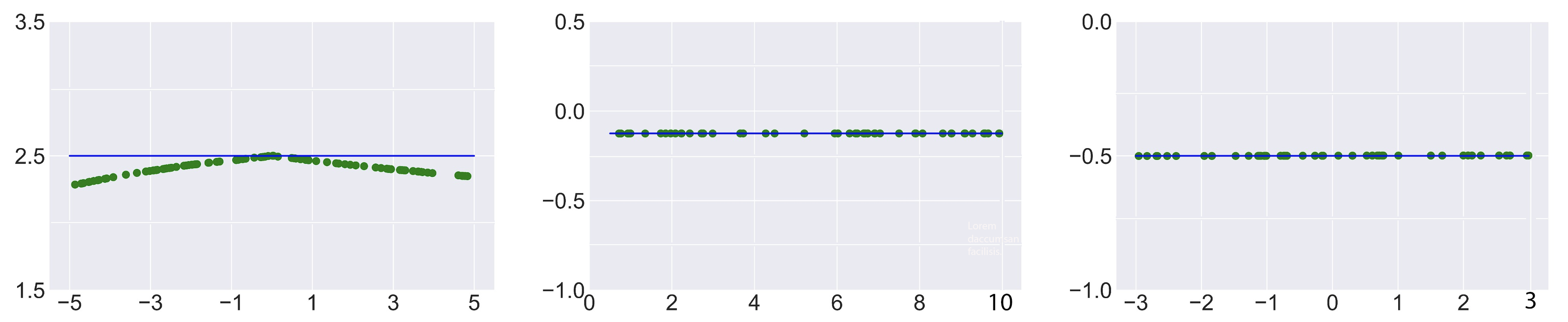} 
        \caption{Position (x-axis) vs Energy (y-axis); Ground (Blue) and Learned (Green Dots) Energies: From left to right: Harmonic Oscillator, Hydrogen Atom (2p case), P\"oschl-Teller (1,1 case).} \label{fig:energy_conservation}
    \end{figure}

In this section we consider some simple one--dimensional time--independent systems. Wave--functions, potential energy and energy levels can be found in Table~\ref{tab:example_wave_function}.  We report our learned potentials in figure~\ref{fig:TDSE} and show that our models obey energy conservation laws in figure~\ref{fig:energy_conservation}. The quantitative results for the experiments can be found in table~\ref{tab:results_supp}. For the derivation of these wave--functions and the general properties of these systems, please see~\cite{Sakurai,feynman1965feynman,robinett}. As before, $\hbar$, $m$ and $\omega$ were set equal to $1$.
\vspace{.1cm}

 \textbf{Quantum Harmonic Oscillator:} The wave--functions in this case are given by Hermite polynomials $H_n, n=0,1,\cdots$. We choose $x \in [-5,5]$ as input to our model. Since $U_{\theta}$ is given by a differential equation (equation~\ref{eqn:time_independent_Schrodinger_loss}) one needs to impose an initial condition to get a unique solution. However, constraining the output of $U_{\theta} \in [0,12.5]$ removes the need for the initial condition. Fig~\ref{fig:TDSE} and fig~\ref{fig:energy_conservation} (left) shows the learned potential and energy of the system.\\
\vspace{.1cm}
\textbf{The Hydrogen Atom (2p case) :}
The general radial wave--functions are given by generalized Laguerre polynomials $L^{l}_{n},  n=1,2,\cdots \ \text{and} \ l=0,1,2,\cdots,n-1$ but in this case simplifies to  $\psi(r) = \frac{1}{8\sqrt{\pi}} re^{\frac{-r}{2}}$. We used  $r \in [0.5,10]$ as input to our model, the initial condition $U(1) = 0.$ and the loss function $\mathcal{L}(\theta) = L(\theta) + U_{\theta}(1)^2$ where $L(\theta)$ is given by equation~\ref{eqn:time_independent_Schrodinger_loss}. Fig~\ref{fig:TDSE} and fig~\ref{fig:energy_conservation} (middle) shows the learned potential and energy of the system.\\
\vspace{.1cm}
\textbf{P\"oschl-Teller potential :} The wave--function $\psi$ generated by this potential is defined by Legendre functions $P_{\lambda}^{\mu}(\text{tanh}(x)),  \lambda =1,2,3\cdot ; \ \mu =1,2,\cdot ,\lambda -1,\lambda$. For simplicity, let $\mu=1$. We choose $x \in [-3,3]$ as input to our model. We imposed an initial condition $U_{\theta}(0) = -\frac{\lambda(\lambda+1)}{2}$ and used a similar auxiliary loss function as above.  Fig~\ref{fig:TDSE} and fig~\ref{fig:energy_conservation} (right) shows the learned potential and energy of the system.

\subsection{Particle in a box (perturbed by some external potential)} 
\begin{figure}[ht]
     \centering
        \includegraphics[width= \linewidth]{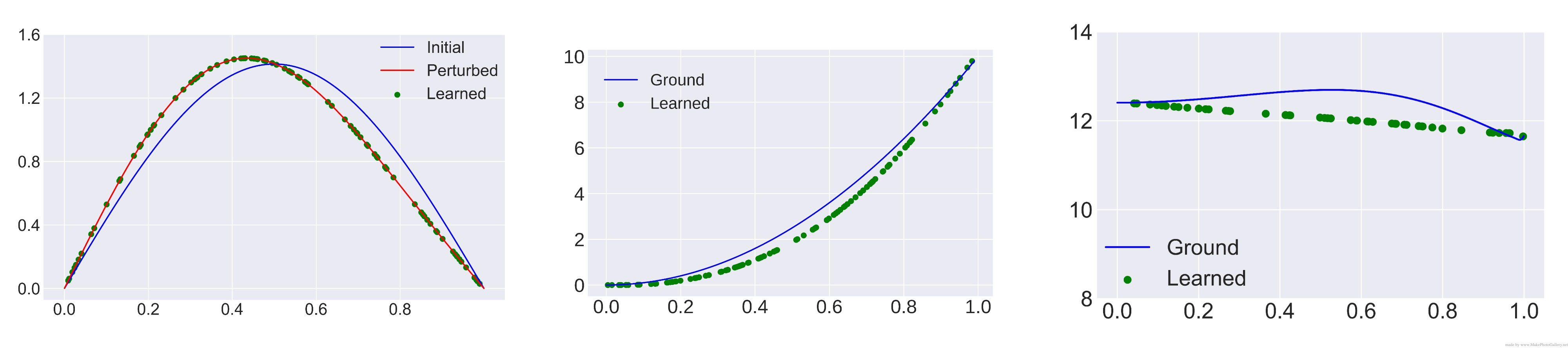} 
         \caption{From left to right: Wave--function of the particle in a box, Potential (y-axis) vs Position (x-axis) of the particle, Conservation of an "energy" like object.} \label{fig:pib}
    \end{figure}

Now, we turn our attention to a quantum system where the Schr\"odinger equation cannot be solved exactly, but can be formulated in an approximate manner using perturbation theory. A particle with no spin, of mass m, was placed in a square one dimensional box, $x\in[0, L]$, of length $L$. Later the particle was presented with the perturbation $V(x)=10x^2$.
The wave--function for the perturbed system was approximated  by considering first order corrections for the unperturbed particle in a box wave--function,
\begin{equation}\label{PIB}
\psi_{n}=\psi^{0}_{n}+\sum_{n\neq k}\dfrac{<\psi^{0}_{n} | V(x) | \psi^{0}_{k}>}{E^{0}_{n}-E^{0}_{k}}\psi^{0}_{k}, \quad n, k=1,2,3,\ldots ,\quad
\end{equation}
where $\psi^{0}_{n}$, and $E^{0}_{n}$ are the unperturbed particle in a box $n$th state wave--function and its energy,  whereas, $<\psi^{0}_{n} | V(x) | \psi^{0}_{k}>$ indicates the following integral
\begin{equation}
<\psi^{0}_{n} | V(x) | \psi^{0}_{k}> =\int{(\psi^{0}_{n})^*V(x)\psi^{0}_{k} dx}.
\end{equation}
For our computations, the wave--function, $\psi^{0}_n$, obtained as solution of the Schr\"{o}edinger equation for the particle in a box model reads,
\begin{equation}
\psi^{0}_{n}=\sqrt{\frac{2}{L}}\text{sin}(\frac{n\pi}{L})x \quad n=1,2,3,\ldots,
\end{equation}
and the energy for the system is given by
\begin{equation}
E^{0}_{n}=\frac{n^2\hbar^{2} \pi^{2}}{2mL^{2}} \quad n=1,2,3,\ldots.
\end{equation}
Here, we use the wave--function only corrected up to a first order for the particle in a box and $x \in [0,1]$. In this experiment we not only learn the potential but also learn the perturbed wave--function based only in the systems initial conditions without perturbation. We use two neural networks, one to learn the potential and the other to learn the perturbed wave--function. The perturbed wave--function was learned in a supervised manner, whereas the potential was learned in an unsupervised manner. If $W_{\theta}$ is the neural network learning the perturbed wave--function $\psi_{\text{pert}}$, then our auxiliary loss function becomes
\begin{equation}
    \mathcal{L}_{\theta} = \big\lvert\big\rvert W_{\theta} - \psi_{\text{pert}} \big\lvert\big\rvert_{2}^{2} + L_{\theta}
\end{equation}
where $L_{\theta}$ is the time--independent Schr\"odinger loss in the main text used to learn the potential and is calculated using the perturbed wave--function.   
Fig~\ref{fig:pib} shows our results on this system. It seems that energy is not conserved for this system but that is merely due to our perturbation approximations.

\subsection{2D Harmonic Oscillator}
\begin{figure}[ht]
     \centering
        \includegraphics[width= \linewidth]{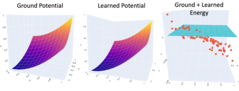} 
         \caption{Potential and Energies of the Harmonic Oscillator} \label{fig:3d_figs_harm}
    \end{figure}

Our work scales easily and quickly to $2$-dimensions as well. The wave--function here is a product of two Hermite polynomials defined above. We choose $x,y \in [0,1]$ as input to our model and constrained our output to $[0,1]$. Thus our loss function is exactly as $1$D Harmonic Oscillator. Fig~\ref{fig:3d_figs_harm} shows our results for this system and the middle figure shows our learned energy is a good approximation to the total energy (z scale chosen from $[4.99,5.01]$).

\section{Motivation behind the time--independent Schr\"{o}dinger Loss}
We present a brief explanation for our time--independent Schr\"{o}dinger Loss function. The Hamilton $\hat{H}$ is the sum of kinetic energy $\hat{T}$ and potential energy $\hat{V}$. The kinetic energy is given by the Laplacian operator. 
\begin{equation}\label{eqn:hamiltonian}
 \hat{H}= -\frac{\hbar^{2}}{2m}\nabla_{x}^{2}+\hat{V}(x).
\end{equation}
For the time independent case, the Schr\"odinger's equation boils down to 
\begin{equation}\label{eqn:energy}
    \hat{H}\psi = E\psi
\end{equation}
where E is the energy of the system. For simplicity, let $\hbar = m =1$.
Using equation~\ref{eqn:hamiltonian}, we can write
\begin{equation}
    (-\frac{1}{2}\nabla_{x}^{2}+\hat{V}(x))\psi = E\psi
\end{equation}
Dividing the above equation by $\psi$ throughout we get 
\begin{equation}\label{eqn:energy_constant}
    \frac{-\frac{1}{2}\nabla_{x}^{2}\psi}{\psi}+\hat{V}(x) = E
\end{equation}
Since the energy is constant, the derivative with respect to $x$ on the left hand side of equation~\ref{eqn:energy_constant} is $0$ and this is our time--independent Schr\"odinger loss function.

\section{Wigner Functions}
An alternative formulation of  quantum dynamics may be given by the Wigner function  \cite{curtright1998features,chen2019numerical}. The Wigner function, $W(x,p,t)$, is a phase space distribution function which behaves similarly to $|\psi\left(x\right)|^2$ and momentum $|\psi\left(p\right)|^2$  distribution functions~\cite{case_Wigner}. Unlike wave--functions, Wigner functions are real valued and bounded. However, contrary to probability distributions,  $W(x,p,t)$ can take negative values. Thus, the Wigner distribution is termed as a quasi--probability distribution and so in a sense loses some of it's classical appeal. Using the Schr\"{o}dinger's equation (equation~\ref{eqn:TDSE}) and the Taylor expansion, the time evolution of the Wigner function is given by an infinite order partial differential equation called Wigner--Moyal equation~\cite{case_Wigner}.
\begin{equation}\label{eqn:Moyal}
\begin{split}
    \frac{\partial W(x,p,t)}{\partial t} &={} -\frac{p}{m} \frac{\partial W(x,p,t)}{\partial x} \\
 {} &  + \sum_{s=0}^{\infty}(-h^2)^{s}\frac{1}{(2s + 1!)}(\frac{1}{2})^{2s} \frac{\partial^{2s+1} U(x)}{\partial x^{2s+1}} \frac{\partial^{2s+1} W(x,p,t)}{\partial p^{2s+1}} 
    \end{split}
\end{equation}

\subsection{Learning Potentials from Wigner Functions}
In the case of the Wigner function, our Neural Network was trained by implementing a truncated Wigner--Moyal loss, 
\begin{dmath}\label{eqn:Moyal-loss}
  L_{Wigner}(\theta) = \bigg\lvert\bigg\rvert\frac{\partial W(x,p,t)}{\partial t}  +\frac{p}{m} \frac{\partial W(x,p,t)}{\partial x}
  - \sum_{s=0}^{k}(-h^2)^{s}\frac{1}{(2s + 1!)}(1/2)^{2s} \frac{\partial^{2s+1} U_{\theta}(x)}{\partial x^{2s+1}} \frac{\partial^{2s+1} W(x,p,t)}{\partial p^{2s+1}} \bigg\lvert\bigg\rvert_{2}^{2} 
  \end{dmath}
where for all our experiments $k=0,1$.  The case where $k=0$ is known as the Liouville equation.
However, we note that equation~\ref{eqn:Moyal-loss} determines $U_{\theta}$ up to a constant. Thus, an initial condition depending on each individual system was added.

\subsection{Experiments with the Wigner functions} 
\subsubsection*{Harmonic Oscillator :}
The Wigner function for the harmonic oscillator has the following form~\cite{case_Wigner}:  
\begin{equation*}
    \begin{aligned}
       W\left(x,p,t\right) &= e^{-(x^2 + p^2)} \left(x^2 + p^2 + \sqrt{2}x\cos{t} - \sqrt{2}p\sin{t} \right)
        \end{aligned}
    \end{equation*}
Since $\frac{\partial^n U}{\partial x^n} = 0, \ \forall \ n \geq 3$, the Moyal--Wigner equation in this case degenerates to the classical Liouville equation. Let $x,p,t \in [0,1]$ and $x$ is the input to the model. The initial condition is $U_{\theta}(0)=0$ and our loss function $\mathcal{L}(\theta) = L_{\text{Wigner}}(\theta) + U_{\theta}(0)^2$ where $L_{\text{Wigner}}$ is given by equation~\ref{eqn:Moyal-loss}. Fig~\ref{fig:Wigner}(left) shows the potential learned by the model.

\subsubsection*{P\"oschl-Teller potential :} The Wigner function in this case~\cite{Chen_2018} is given by:
\begin{dmath}\label{xx}
W_{2,1,0}(x,k,t) := \frac{3}{8}\int^{\infty}_{-\infty} \text{sech}^2(x + \frac{y}{2})\text{sech}^2(x - \frac{y}{2})
     \times 
      \bigg[2\text{sinh}(x + \frac{y}{2}) \text{sinh}(x - \frac{y}{2}) +
\sqrt{2}\text{sinh}(x - \frac{y}{2})e^{\frac{i3t}{2}} +
\sqrt{2} \text{sinh}(x +\frac{y}{2})e^{\frac{-i3t}{2}} + 1\bigg]e^{-iky}dy 
\end{dmath}

The Wigner function is a real--valued bounded function. Thus by breaking the  integral in equation~\ref{xx} into real and complex parts, we only focus on the real part. Using Euler's formula, we get the following: 
\begin{dmath}\label{eqn:real_part}
g_{2,1,0}(x,k,t) =  \frac{3}{8}\int_{-\infty}^{\infty} \text{sech}^2(x + \frac{y}{2})\text{sech}^2(x - \frac{y}{2}) \times \bigg[2\text{sinh}(x + \frac{y}{2}) \text{sinh}(x - \frac{y}{2})\text{cos}(-ky) + \sqrt{2}\text{sinh}(x - \frac{y}{2})\text{cos}(\frac{3t}{2}-ky) + \sqrt{2}\text{sinh}(x + \frac{y}{2})\text{cos}(-\frac{3t}{2}-ky)+\text{cos}(-ky) \bigg]
\end{dmath}

Note that the integral in equation~\ref{eqn:real_part} is invariant under the change of variable $y \rightarrow -y$. This implies in order to calculate $g_{2,1,0}(x,k,t)$, we only have to integrate from $0$ to $\infty$ and multiply that integral by 2. Our final simplification comes from studying the decay properties of the Wigner functions. Using $\text{sech}(x) = \frac{2}{e^x + e^{-x}}$ and $\text{sinh}(x) = \frac{e^{x} - e^{-x}}{2}$, we found that the integrand in equation~\ref{eqn:real_part} behaves like $O(e^{-y})$ (resp. $O(e^{y})$) as $y \rightarrow \infty$ (resp. $y \rightarrow -\infty$). We picked a threshold of $10^{-9}$ to truncate the integral from positive real axis to a bounded interval which gives the following form :
\begin{dmath}\label{yy}
f_{2,1,0}(x,k,t) = \frac{3}{4}\int_{0}^{20} \text{sech}^2(x + \frac{y}{2})\text{sech}^2(x - \frac{y}{2}) \times \bigg[2\text{sinh}(x + \frac{y}{2}) \text{sinh}(x - \frac{y}{2})\text{cos}(-ky) + \sqrt{2}\text{sinh}(x - \frac{y}{2})\text{cos}(\frac{3t}{2}-k) + \sqrt{2}\text{sinh}(x + \frac{y}{2})\text{cos}(-\frac{3t}{2}-ky)+\text{cos}(-ky) \bigg]
\end{dmath}

\begin{figure}[ht]
    \centering
        \includegraphics[width=\linewidth]{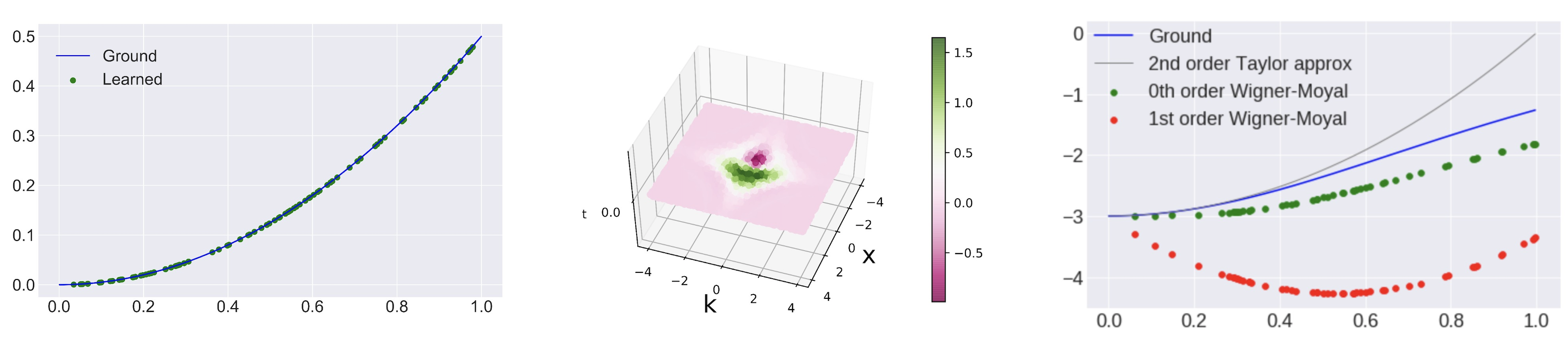} 
      \caption{Left to right: Position (x-axis) vs Potential (y-axis) of Harmonic oscillator using the Wigner function; Our approximation of the Wigner function of P\"oschl--Teller potential; Various approximations of the potential using the Wigner--Moyal equation.} \label{fig:Wigner}
    \end{figure}

The Wigner method to study the time--frequency properties of dynamical systems involves taking the partial derivatives with respect to time  of the Wigner function. These derivatives on the Wigner function yield what is known as the Wigner--Moyal equation. The  physical interpretations, numerical difficulties and approximations of the Wigner--Moyal equation have been widely discussed  in the literature, thus for  information about the  mathematical challenges associated with the Wigner--Moyal equation, we recommend readers to consult these references ~\cite{Chen_2018,case2008wigner,galleani2002approximation,heller1976wigner, curtright1998features,klimov2020truncated,athanassoulis2008exact,gomes2008wigner}.\\
The potential is $U(x) = -3\text{sech}(x)^2$ which is infinitely differentiable. We choose $x \in [0,1]$ as input to our model.  In this experiment we will attempt to approximate the infinite order PDE (equation~\ref{eqn:Moyal}) by equation~\ref{eqn:Moyal-loss}. 
One then cannot assume that any non--steady state solution predicted by the truncated Wigner function is immediately valid, as it can be shown that higher order quantum corrections are responsible for quantum mechanical phase space behavior. The $0$th order truncation matches the potential in a small neighborhood of $0$. Figure~\ref{fig:Wigner} summarizes some of our findings, for more details on approximation procedures and their challenges for the infinite order Wigner--Moyal PDE see supplementary material.

\begin{table*}[t]
\renewcommand{\arraystretch}{0.5}
\begin{center}
\caption{A quantitative analysis of our model }
\label{tab:results_supp}
 \resizebox{\textwidth}{!}{\begin{tabular}{|l c c |} 
 \hline
 System & RMSE between True and Learned Potentials & RMSE between True and Learned Energies \\ [0.5ex] 
 \hline\hline
 Harmonic Oscillator& $\num{1.1e-1} \pm \num{5.0e-2}$&  $\num{1.0e-1} \pm \num{2.0e-2}$\\[0.9ex] 
 P\"ochl--Teller potential& $\num{1.0e-4} \pm \num{6.0e-5}$& $\num{8.0e-4} \pm \num{6.0e-5}$\\[0.9ex]
Radial Hydrogen atom& $\num{3.0e-4} \pm \num{8.0e-5}$ & $\num{3.0e-4} \pm \num{7.0e-5}$\\[0.9ex]
 2D Harmonic Oscillator& $\num{3.0e-3} \pm \num{9.0e-4}$ & $ \num{4.0e-3}\pm \num{8.0e-4}$\\[0.9ex] 
 
 Particle in a Box& $\num{4.3e-1} \pm\num{6.0e-2}$ & $\num{5.5e-1} \pm \num{8.0e-1}$\\[0.9ex]
 
 Soliton & $\num{2.9e-1} \pm \num{4.0e-2}$& -\\[0.9ex]
 
 Harmonic Oscillator from Wigner & $\num{4.0e-3}\pm \num{8.0e-5}$& -\\[.09ex]
 \hline
\end{tabular}}
\end{center}
\end{table*}

\section{Some training details and hyperparameters}
Our Neural Network is a 4-layer feedforward network with a residual connection between the second and the third layers. The activation and the scaling in the final layers varied from experiment to experiment. Our main motivation for scaling and using different activation is to show that an appropriate architecture can perfectly learn the correct potential without an initial condition. All the models are trained for $1000$ epochs. Table~\ref{tab:model} shows the activation, scaling and the size of the training data for each of the studied systems. All the training data was randomly sampled from the appropriate domains and  trained in a minibatch fashion with batch size $32$.

\begin{table*}[t]
\renewcommand{\arraystretch}{0.5}
\begin{center}
\caption{Some training details and model hyperparameters}
\label{tab:model}
 \resizebox{\textwidth}{!}{\begin{tabular}{l c c c } 
 \hline
 System & Final Layer Activation & Final Layer Scaling & Size of training data \\ [0.5ex] 
 \hline\hline
 Harmonic Oscillator & Sigmoid &  $12.5$ & $2500$\\[0.9ex] 
 P\"ochl--Teller potential & None & None & $2500$\\[0.9ex]
Radial Hydrogen atom& None & None & $2500$\\[0.9ex]
 2D Harmonic Oscillator& Sigmoid & None & $5000$\\[0.9ex] 
 
 Potential for Particle in a Box  & Sigmoid & 10 & $4000$\\[0.9ex]
 
 Perturbation for Particle in a Box  & None & None & $4000$\\[0.9ex]
 
 Soliton & None & None & $3000$\\[0.9ex]
 
 Harmonic Oscillator from Wigner & None & Sigmoid & $5000$\\[.9ex]
 
 P\"oschl--Teller from Wigner & None & None & $2000$ \\
 \hline
\end{tabular}}
\end{center}
\end{table*}

\section{Additional Figures}
Some additional figures of the wave--functions and Wigner functions used in our experiments. 
\begin{figure}[ht]
    \centering
        \includegraphics[width=\linewidth]{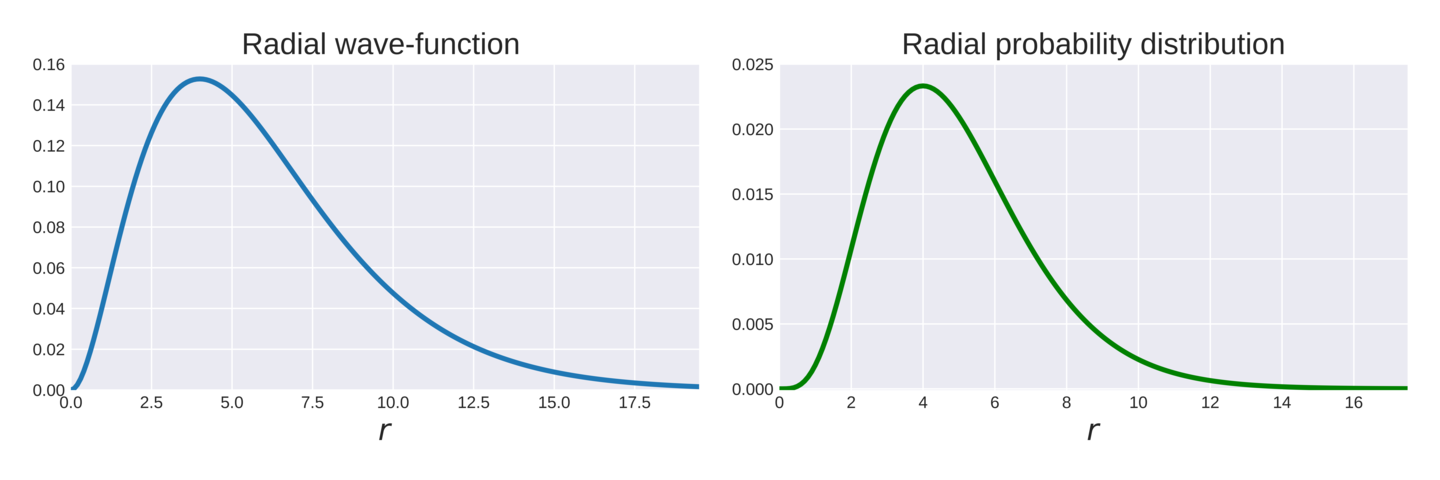} 
      \caption{Left: $2p$ Radial wave--function for the Hydrogen atom. Right: $2p$ Radial probability distribution for the Hydrogen atom } \label{fig:Hatom}
    \end{figure}

\begin{figure}[ht]
    \centering
        \includegraphics[width=\linewidth]{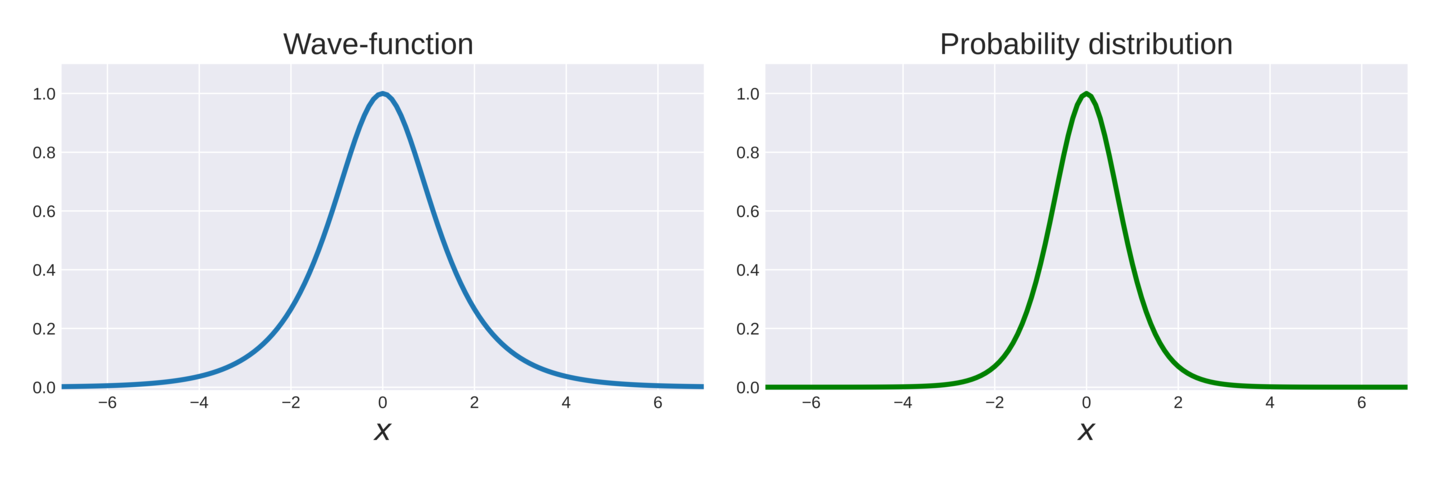} 
      \caption{ Wave--function (Left) and Probability distribution (Right)  for the first bound state of the P\"{o}schl--Teller potential.} \label{fig:Poschl-Teller_potential}
    \end{figure}
 
\begin{figure}[ht]
    \centering
        \includegraphics[width=.47\linewidth]{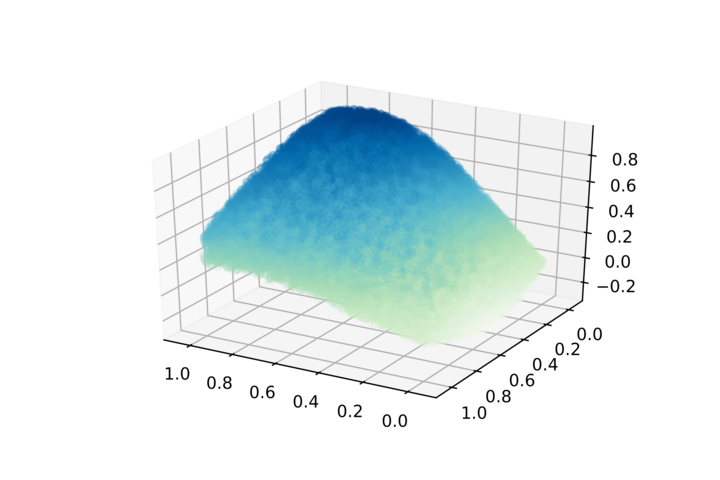} 
        \includegraphics[width=.47\linewidth]{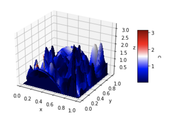} 
        
      \caption{Left: Wigner quasi--probability distribution for the ground state of the Harmonic Oscillator, Right: Our approximation of the Wigner function for the P\"oschl--Teller  } \label{fig:extra_wigner}
    \end{figure}

\begin{figure}[t]
    \centering
        \includegraphics[width=.47\linewidth]{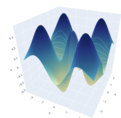} 
        \includegraphics[width=.47\linewidth]{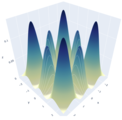} 
        
      \caption{Left: 2D Harmonic Oscillator Wave--function, Right: 2D Harmonic Oscillator Wave--function } \label{fig:extra_2d}
    \end{figure}

\end{document}